\documentclass{article}

\usepackage{spconf,amsmath,graphicx}
\usepackage{booktabs,multirow}
\usepackage{amssymb}
\usepackage{hyperref}
\usepackage{cleveref}
\usepackage{setspace}

\title{VGDiffZero: Text-to-image Diffusion Models Can Be Zero-shot Visual Grounders}
\name{Xuyang Liu$^{1,2*}$
Siteng Huang$^{2*}$
Yachen Kang$^2$
Honggang Chen$^1$
Donglin Wang$^{2\dag}$
\thanks{$^*$Equal contribution. $^\dag$Corresponding author.}}
\address{$^1$College of Electronics and Information Engineering, Sichuan University, Chengdu, China\\
$^2$School of Engineering, Westlake University, Hangzhou, China}
\begin{document}

%
\maketitle
\begin{abstract}
Large-scale text-to-image diffusion models have shown impressive capabilities for generative tasks by leveraging strong vision-language alignment from pre-training. However, most vision-language discriminative tasks require extensive fine-tuning on carefully-labeled datasets to acquire such alignment, with great cost in time and computing resources. In this work, we explore directly applying a pre-trained generative diffusion model to the challenging discriminative task of visual grounding without any fine-tuning and additional training dataset. Specifically, we propose VGDiffZero, a simple yet effective zero-shot visual grounding framework based on text-to-image diffusion models. We also design a comprehensive region-scoring method considering both global and local contexts of each isolated proposal. Extensive experiments on RefCOCO, RefCOCO+, and RefCOCOg show that VGDiffZero achieves strong performance on zero-shot visual grounding. Our code is available at \url{https://github.com/xuyang-liu16/VGDiffZero}.
\end{abstract}
\section{Introduction}
\label{sec:intro}

Recently, large-scale vision-language pre-trained models \cite{radford2021learning,openai2023gpt4} have demonstrated strong performance across a wide range of downstream tasks. Among them, most vision-language (VL) tasks can be broadly categorized into two types - generative tasks and discriminative tasks. For VL generative tasks, text-to-image generative diffusion models, such as Stable Diffusion \cite{rombach2022high}, have shown their powerful ability to generate high-fidelity and diverse images from text descriptions. To be specific, these generative diffusion models are firstly pre-trained on large-scale text-image pairs datasets like LAION-5B \cite{schuhmann2022laion5b} to learn the correlations between textual descriptions and visual concepts, then applying them in VL generative tasks, such as image editing \cite{kawar2023imagic} and inpainting \cite{xie2023smartbrush}. For VL discriminative tasks, the predominant approaches are to obtain vision-language alignment abilities from large-scale pre-training datasets, which can then be utilized to facilitate downstream visual-language reasoning tasks \cite{zhang2021vinvl,liu2023dap,liu2023prompt}. Among various VL discriminative tasks, visual grounding \cite{yu2016modeling} is one of the most challenging one, in which the aim is to localize a target object in an image given a textual description. Most supervised visual grounding works \cite{yu2018mattnet,su2023language,liu2023dq} study how to effectively fuse cross-modality features extracted independently by each encoder. Though these methods have achieved good performance on general visual grounding benchmarks, such as RefCOCO and RefCOCO+ \cite{yu2016modeling}, the training and fine-tuning cost is prohibitively expensive and time-consuming. Moreover, collecting task-specific data is even more challenging since it requires accurate descriptions of the target area as well as high-quality bounding box annotations \cite{yu2023zero}.

\begin{figure}[t]
\centering
\includegraphics[width=0.48\textwidth]{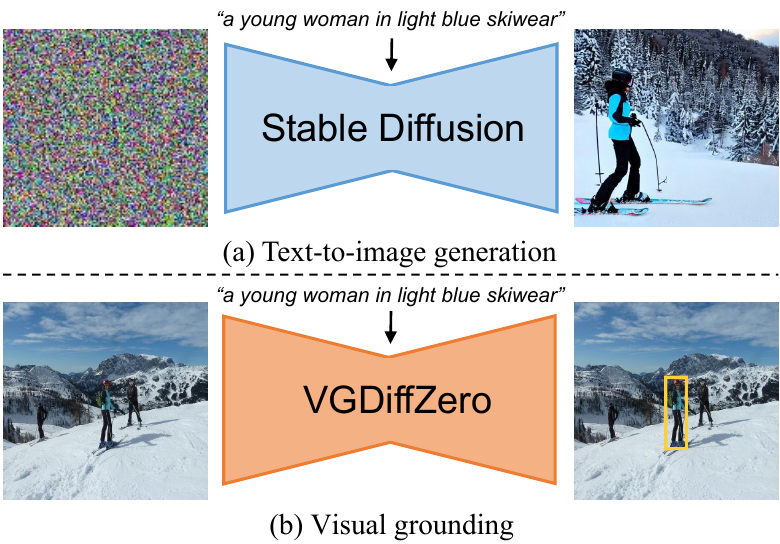}
\vspace{-7mm}
\caption{Illustration of two types of vision-language tasks. Motivated by the strong abilities of text-to-image diffusion models, we propose VGDiffZero for zero-shot visual grounding.}
\vspace{-3mm}
\label{fig1}
\end{figure}

Recent studies start to leverage pre-trained diffusion models for a set of discriminative tasks, including classification \cite{li2023your}, segmentation \cite{zhao2023unleashing,karazija2023diffusion} and image-text matching \cite{he2023discriminative}, under different settings. These progresses fully demonstrate \textbf{two key advantages} of text-to-image diffusion models: (1) Strong abilities of vision-language alignment. (2) Sufficient knowledge of spatial relations and fine-grained disentangled concepts. These above two advantages inspire an intriguing question: \textit{is it possible to directly adapt these advantages of pre-trained text-to-image diffusion models to visual grounding, without costly fine-tuning?}

Motivated by the above observations, in this paper, we seek to directly leverage the power of pre-trained generative diffusion models, particularly Stable Diffusion \cite{rombach2022high}, for a VL discriminative task of visual grounding, as shown in \Cref{fig1}. Specifically, we propose \textbf{VGDiffZero}, a simple yet novel zero-shot visual grounding framework using text-to-image diffusion models. Overall, VGDiffZero regards the visual grounding task as a isolated proposal selection process mainly through our proposed comprehensive region-scoring method including two stages: \textbf{Noise Injection} and \textbf{Noise Prediction}. In the first stage, a series of detected object proposals undergo masking and cropping to obtain isolated proposals with global and local visual information. These isolated proposals are then encoded into latent vectors, and injected with Gaussian noise. In the second stage, each noised latent vector along with the text embeddings encoded by a pre-trained CLIP text encoder, is fed into the denoisng UNet to predict the injected noise. By comparing all predicted and sampled noise pairs, the best matching proposal with the minimum errors is output as the final prediction.

In general, our main contributions can be summarized as threefold: (1) We propose a novel diffusion-based framework, termed VGDiffZero, for zero-shot visual grounding without any additional fine-tuning. To the best of our knowledge, this is the first attempt to tackle visual grounding using a generative diffusion model under the zero-shot setting. (2) We propose a comprehensive region-scoring method that incorporates global and local contexts of input images to enable accurate proposal selection. (3) Extensive experiments on visual grounding benchmarks of RefCOCO \cite{yu2016modeling}, RefCOCO+ \cite{yu2016modeling} and RefCOCOg \cite{mao2016generation} demonstrate the effectiveness of our proposed VGDiffZero.

\section{Methodology}
\label{sec:methodology}

In this section, we introduce VGDiffZero, our proposed diffusion-based framework for zero-shot visual grounding. We first revisit Stable Diffusion, the foundation of our model, and then delineate the two key stages in VGDiffZero.

\begin{figure*}[h]
\centering
\vspace{-3mm}
\includegraphics[width=\textwidth]{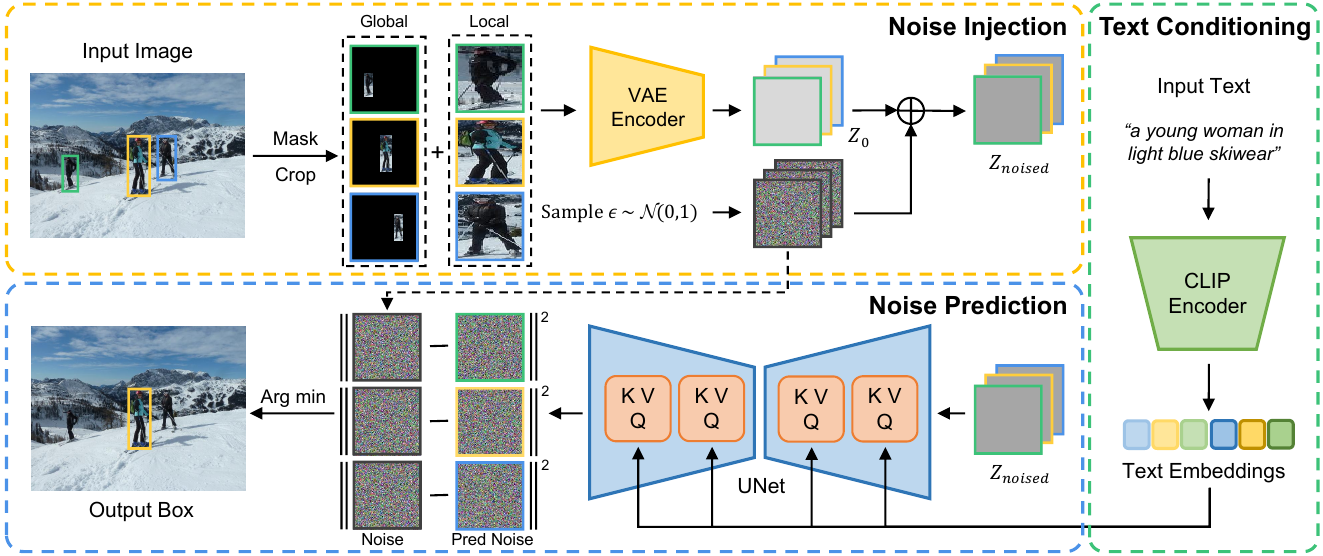}
\vspace{-7mm}
\caption{Overview of our VGDiffZero. Given an input image, isolated proposals are generated via cropping and masking, and then encoded individually into latent vectors $Z_0$. Gaussian noise $\epsilon$ sampled from $\mathcal{N}(0, 1)$ is injected into each latent vector to obtain noised latent representations $Z_{noised}$. Subsequently, each noised latent together with the text embeddings is fed into the UNet to select the best matching proposal as the final prediction.}
\vspace{-3mm}
\label{fig2}
\end{figure*}

\subsection{Preliminaries: Stable Diffusion}
\label{sec:preliminaries}

Diffusion models \cite{sohl2015deep, ho2020denoising, rombach2022high} represent a novel class of generative models that train neural networks to reverse a deterministic diffusion process. Stable Diffusion \cite{rombach2022high} is a kind of latent diffusion models which implements diffusion process in the latent space, rather than data space. Specifically, Stable Diffusion consists of three components: a Variational Autoencoder (VAE) including encoder and decoder, a diffusion model including the denoising UNet \cite{ronneberger2015u} and DDPM Sampler \cite{ho2020denoising}, and a text encoder of the pre-trained CLIP \cite{radford2021learning}.

During training, Stable Diffusion learns to invert the latent diffusion process over image-text pairs $(x,y)$. To be specific, the VAE encoder first maps an image $x$ to latent vectors $z$, and Gaussian noise $\epsilon \sim \mathcal{N}(0,I)$ is then iteratively added to $z$:
\begin{equation}
q(z_t|z_{t-1}) = \mathcal{N}(z_t; \sqrt{1-\beta_t}z_{t-1}, \beta_t I), t = 1,...,T,
\end{equation}
where $q(z_t|z_{t-1})$ is the conditional density of $z_t$ given $z_{t-1}$, $\left(\beta_t \right)^T_{t=1}$ are hyperparameters that determine the noise schedule, and $T$ is the total timesteps. The denoising UNet takes $z_t$, current timestep $t$, and text embeddings $\tau_\theta(y)$ as inputs to predict the noise $\epsilon_t$ as:
\begin{equation}
\epsilon_t = \text{UNet}(z_t, t, \tau_\theta(y)),
\end{equation}
where the text embeddings $\tau_\theta(y)$ is obtained from text $y$ via CLIP text encoder. The training target is to minimize the predicted error $e$ as:
\begin{equation}
e = ||\epsilon - \epsilon_t(z_t, t, \tau_\theta(y))||^2,
\end{equation}
where $\epsilon$ and $\epsilon_t$ respectively represent sampled Gaussian noise and the predicted noise by the denoising UNet.

During generation, input text $y$ is first encoded into text embeddings $\tau_\theta(y)$ via the text encoder of CLIP. The latent vector $z_T$ is sampled from the standard normal distribution $\mathcal{N}(0,I)$, and denoising UNet takes $z_T$ and $\tau_\theta(y)$ to remove the noise to recover the denoised latent vector $z_0$ recursively. Finally, the denoised latent vector $z_0$ is decoded into an image via the VAE decoder. In this way, Stable Diffusion is able to generate images conditioned on the input text.

\subsection{Zero-shot Visual Grounding via Diffusion Models}
\label{sec:method}

Generally, the visual grounding task can be viewed as the process of selecting the proposal that best fits the textual query \cite{yu2018mattnet,yao2021cpt}. This involves two critical aspects: (1) Generating high-quality isolated proposals, by taking into account comprehensive visual information of each proposal, including the location within the global image as well as the internal visual information within the local proposal. (2) Establishing fine-grained alignments between region proposals and the textual query, by leveraging the vision-language matching abilities from large-scale pre-training datasets. To this end, we adopt VGDiffZero equipped with a designed comprehensive region-scoring method for zero-shot visual grounding. As depicted in \Cref{fig2}, VGDiffZero consists of two key stages: Noise Injection and Noise Prediction.

\textbf{Noise Injection.} Given that visual grounding involves identifying the region proposal best aligned with the textual query, the first step is to generate multiple object proposals from the input image. In this case, following the previous works \cite{yu2018mattnet,yao2021cpt}, we adopt a pre-trained object detector, \textit{i.e.,} Faster R-CNN \cite{ren2015faster} to extract potential region representations in the image. In order to isolate proposals preserving their locations within the whole image and internal visual information, we devise a comprehensive approach that involves \textbf{masking} out all image except for the proposal region, alongside \textbf{cropping} to only retain the proposal region. After processing all proposals via the comprehensive isolation approach, we acquire two sets of isolated proposals, including the global set and local set (corresponding to Global and Local in \Cref{fig2}, respectively). These are then individually encoded by the VAE encoder to obtain the latent representations $Z_0$ of each proposal. Gaussian noise $\epsilon \sim \mathcal{N}(0,I)$ is injected into each latent vector to produce the noised latent representations $Z_{noised}$. In summary, the Noise Injection process accomplishes three objectives: isolating region proposals, encoding proposals into latent representations, and injecting noise to diffuse the latents forward.

\textbf{Noise Prediction}. Given an input text $y$, it is first encoded by the pre-trained CLIP text encoder to derive the text embeddings $\tau_\theta(y)$. The denoising UNet takes two sets of noised latent vectors $Z_\text{noised}$ and text embeddings $\tau_\theta(y)$ to predict the sampled noise $\epsilon$ for each isolated proposal. Subsequently, two sets of prediction errors, $e_\text{mask}$ and $e_\text{crop}$, are computed for each proposal by calculating the deviation between the predicted noise and sampled noise (corresponding to Pred Noise and Noise in \Cref{fig2}, respectively). Since Stable Diffusion is pre-trained on semantic-consistent image-text pairs, a smaller error indicates the model more accurately predicts the noise conditioned on $z_t$ and $\tau_\theta(y)$, meaning the current region and text are more semantically aligned. Finally, we compute the total error $e_\text{total} = e_\text{mask} + e_\text{crop}$ for each proposal and select the proposal with the minimum total error $e_\text{total}$ as the prediction output. In this way, our proposed VGDiffZero can consider both the global and local contexts of each isolated proposal for comprehensive proposal selection.

\section{Experiments}
\label{sec:pagestyle}

\begin{table*}[t]
\centering
\vspace{-3mm}

\begin{tabular}{lcccccccccc}
\toprule

\multirow{2}{*}{Methods} & \multicolumn{3}{c}{RefCOCO} & & \multicolumn{3}{c}{RefCOCO+} & & \multicolumn{2}{c}{RefCOCOg} \\
\cline{2-4}
\cline{6-8}
\cline{10-11}
& val & test A & test B & & val & test A & test B & & val & test \\
\midrule

Random & 15.61 & 13.47 & 18.23 & & 16.30 & 13.29 & 19.98 & & 18.79 & 18.35 \\
CPT-Blk & 26.90 & 27.50 & 27.40 & & 25.40 & 25.00 & 27.00 & & 32.10 & 32.30 \\
Cropping & 26.04 & 26.34 & 28.95 & & 26.34 & 26.28 & 29.41 & & 31.64 & 32.37 \\
Masking & 27.17 & 29.47 & 26.21 & & 27.64 & 29.62 & 27.29 & & 32.66 & 32.56 \\
VGDiffZero w/ Single IPM & 26.78 & 29.56 & 27.28 & & 27.41 & 29.55 & 27.21 & & 32.82 & 32.39 \\
\textbf{VGDiffZero}	& \textbf{27.95} & \textbf{30.34} & \textbf{29.11} & & \textbf{28.39} & \textbf{30.79} & \textbf{29.79} & & \textbf{33.53} & \textbf{33.24} \\

\bottomrule
\end{tabular}
\vspace{-2mm}
\caption{Comparison of accuracy (\%) on RefCOCO \cite{yu2016modeling}, RefCOCO+ \cite{yu2016modeling} and RefCOCOg \cite{mao2016generation} under the zero-shot setting.} 
\vspace{-3mm}
\label{tab1}
\end{table*}

\subsection{Experimental Settings}
\label{sec:experiments}

\textbf{VGDiffZero Setup.} We use the detected object proposals from a pre-trained Faster R-CNN \cite{ren2015faster}, and each isolated proposal (both masking and cropping) is resized to $512\times512$. VGDiffZero is built on the pre-trained Stable Diffusion 2.1-base ~\cite{rombach2022high} with DDPM Sampler \cite{ho2020denoising} and 1,000 timesteps. We use the text encoder initialized from CLIP-ViT-H/14 \cite{radford2021learning}.

\noindent \textbf{Datasets and Evaluation Metrics.} We evaluate VGDiffZero on three widely-used VG benchmarks: RefCOCO \cite{yu2016modeling}, RefCOCO+ \cite{yu2016modeling} and RefCOCOg \cite{mao2016generation}. RefCOCO, RefCOCO+, and RefCOCOg datasets contain 19,994, 19,992, and 26,771 images with 142,210, 141,564, and 104,560 referring expressions, respectively. RefCOCO and RefCOCO+ have shorter expressions (avg 1.6 nouns, 3.6 words), while RefCOCOg has longer, more complex expressions (avg 2.8 nouns, 8.4 words). We use Accuracy@0.5 as the evaluation metrics.

\noindent \textbf{Baselines.} We compare VGDiffZero with related methods: (1) Random: Randomly selecting a detected proposal as the prediction. (2) CPT-Blk \cite{yao2021cpt}: A strong zero-shot visual grounding baseline that shades detected proposals with different colors and uses a masked language prompt in which the referring expression is followed by “in \texttt{[MASK]} color". The color with the highest probability by the pre-trained masked language model VinVL \cite{zhang2021vinvl} is chosen as prediction. (3) Cropping: Isolating detected proposals by cropping the image around each proposal, then passing the cropped images through VGDiffZero. (4) Masking: Similar to Cropping, but using masking to isolate detected proposals before passing through VGDiffZero. (5) VGDiffZero w/ Single IPM: Given two sets of prediction errors from isolating proposal methods (IPM), cropping and masking, selecting the proposal with the lowest prediction error as the final prediction.

\subsection{Experimental Results and Analysis}
\label{sec:results}

\textbf{Quantitative Comparison.} The main experimental results are reported in \Cref{tab1}, from which we can observe that: (1) Our proposed VGDiffZero outperforms other zero-shot visual grounding baseline methods. Notably, on the RefCOCO+ test A, VGDiffZero achieves 5.79\% higher accuracy compared to CPT-Blk \cite{yao2021cpt}. This indicates that VGDiffZero can effectively and directly leverage the vision-language alignment abilities learned by pre-trained VL generative models to perform well on visual grounding task. (2) Masking to isolate object proposals achieves superior performance compared to cropping on most datasets, which suggests that preserving the global location of proposals plays a role in the visual grounding task. (3) Using both masking and cropping to isolate proposals achieves higher accuracy than using either method alone. This indicates that considering both the global and local contexts of isolated proposals enables more robust performance on visual grounding tasks. 

\begin{table}[t]
\centering

\begin{tabular}{lccc}
\toprule
Methods & RefCOCO & RefCOCO+ & RefCOCOg \\
\midrule

w/ core-exp & 26.86 & 27.13 & \textbf{34.32} \\
\textbf{w/ full-exp} & \textbf{27.95} & \textbf{28.39} & 33.53 \\

\bottomrule
\end{tabular}
\vspace{-2mm}
\caption{Accuracy on the validation sets of RefCOCO, RefCOCO+ and RefCOCOg given the core expression and full expression, respectively.} 
\vspace{-3mm}
\label{tab2}
\end{table}

\noindent \textbf{Effect of Different Expression Processing Methods.} To further investigate the impact of different expression processing methods on diffusion-based visual grounding, we compare two expression processing approaches: using the full expression as input to the text encoder of CLIP versus using core expression extracted by spaCy \cite{honnibal2015improved}. As summarized in \Cref{tab2}, using the full expression as text input performs better on the validation sets of RefCOCO and RefCOCO+, suggesting that preserving the complete expression is more advantageous when given expressions are short and contain few objects, such as \textit{“A young woman in lightblue skiwear"}. In contrast, extracting the core noun phrase is more suitable for handling complex sentences with multiple objects, such as \textit{“A little rabbit crouching in the bushes under the shade of a tree"}, since spaCy can extract the core expression \textit{“A little rabbit"} to better separate the object from its surroundings.

\begin{table}[t]
\centering

\begin{tabular}{cccc}
\toprule

SD Ver. & RefCOCO & RefCOCO+ & RefCOCOg \\
\midrule

1-2 & 27.11 & 26.73 & 32.34 \\
1-4 & 27.64 & 27.61 & 32.73 \\
1-5 & 27.86 & 27.97 & 32.81 \\
\textbf{2-1} & \textbf{27.95} & \textbf{28.39} & \textbf{33.53} \\

\bottomrule
\end{tabular}
\vspace{-2mm}
\caption{Accuracy on the validation sets of RefCOCO, RefCOCO+ and RefCOCOg using different versions of Stable Diffusion (short for SD Ver.).}
\vspace{-3mm}
\label{tab3}
\end{table}

\noindent \textbf{Effect of Different Pre-trained Models.} Since our VGDiffZero is built on pre-tained Stable Diffusion \cite{rombach2022high}, it is necessary to investigate how the pre-trained models impacts the visual grounding performance. We compare four versions of pre-trained Stable Diffusion and summarize the results on three benchmark validation sets in \Cref{tab3}. It is clear that from SD-1-2 to SD-2-1, the accuracy on three benchmarks are improved, which suggests that larger-scale pre-training can improve the vision-language alignment abilities of text-to-image diffusion models, which are precisely needed for visual grounding tasks, thus enabling superior performance.

\section{Conclusions}
\label{sec:typestyle}
In this paper, we propose VGDiffZero, a novel zero-shot visual grounding framework that leverages pre-trained text-to-image diffusion models' vision-language alignment abilities. Through the designed comprehensive region-scoring method, our VGDiffZero can consider both the global and local contexts of each isolated object proposal. Extensive experimental results demonstrate that VGDiffZero achieves satisfactory performance on three general visual grounding benchmarks. 

\noindent \textbf{Acknowledgements.} This research was supported by STI 2030—Major Projects (2022ZD0208800), NSFC General Program (Grant No. 62176215), and NSFC Young Scientists Fund (Grant No. 62001316).


\bibliographystyle{IEEEbib}
\bibliography{strings,refs}

\begin{thebibliography}{10}

\bibitem{radford2021learning}
Alec Radford, Jong~Wook Kim, Chris Hallacy, Aditya Ramesh, Gabriel Goh,
  Sandhini Agarwal, Girish Sastry, Amanda Askell, Pamela Mishkin, Jack Clark,
  et~al.,
\newblock ``Learning transferable visual models from natural language
  supervision,''
\newblock in {\em ICML}, 2021.

\bibitem{openai2023gpt4}
OpenAI,
\newblock ``{GPT-4} technical report,'' 2023.

\bibitem{rombach2022high}
Robin Rombach, Andreas Blattmann, Dominik Lorenz, Patrick Esser, and Bj{\"o}rn
  Ommer,
\newblock ``High-resolution image synthesis with latent diffusion models,''
\newblock in {\em CVPR}, 2022.

\bibitem{schuhmann2022laion5b}
Christoph Schuhmann, Romain Beaumont, Richard Vencu, Cade Gordon, Ross
  Wightman, Mehdi Cherti, Theo Coombes, Aarush Katta, Clayton Mullis, Mitchell
  Wortsman, et~al.,
\newblock ``{LAION-5B}: An open large-scale dataset for training next
  generation image-text models,''
\newblock {\em arXiv preprint arXiv:2210.08402}, 2022.

\bibitem{kawar2023imagic}
Bahjat Kawar, Shiran Zada, Oran Lang, Omer Tov, Huiwen Chang, Tali Dekel, Inbar
  Mosseri, and Michal Irani,
\newblock ``Imagic: Text-based real image editing with diffusion models,''
\newblock in {\em CVPR}, 2023.

\bibitem{xie2023smartbrush}
Shaoan Xie, Zhifei Zhang, Zhe Lin, Tobias Hinz, and Kun Zhang,
\newblock ``{SmartBrush}: Text and shape guided object inpainting with
  diffusion model,''
\newblock in {\em CVPR}, 2023.

\bibitem{zhang2021vinvl}
Pengchuan Zhang, Xiujun Li, Xiaowei Hu, Jianwei Yang, Lei Zhang, Lijuan Wang,
  Yejin Choi, and Jianfeng Gao,
\newblock ``{VinVL}: Revisiting visual representations in vision-language
  models,''
\newblock in {\em CVPR}, 2021.

\bibitem{liu2023dap}
Ting Liu, Yue Hu, Wansen Wu, Youkai Wang, Kai Xu, and Quanjun Yin,
\newblock ``Dap: Domain-aware prompt learning for vision-and-language
  navigation,''
\newblock {\em arXiv preprint arXiv:2311.17812}, 2023.

\bibitem{liu2023prompt}
Ting Liu, Wansen Wu, Yue Hu, Youkai Wang, Kai Xu, and Quanjun Yin,
\newblock ``Prompt-based context-and domain-aware pretraining for vision and
  language navigation,''
\newblock {\em arXiv preprint arXiv:2309.03661}, 2023.

\bibitem{yu2016modeling}
Licheng Yu, Patrick Poirson, Shan Yang, Alexander~C Berg, and Tamara~L Berg,
\newblock ``Modeling context in referring expressions,''
\newblock in {\em ECCV}, 2016.

\bibitem{yu2018mattnet}
Licheng Yu, Zhe Lin, Xiaohui Shen, Jimei Yang, Xin Lu, Mohit Bansal, and
  Tamara~L Berg,
\newblock ``{MAttNet}: Modular attention network for referring expression
  comprehension,''
\newblock in {\em CVPR}, 2018.

\bibitem{su2023language}
Wei Su, Peihan Miao, Huanzhang Dou, Gaoang Wang, Liang Qiao, Zheyang Li, and
  Xi~Li,
\newblock ``Language adaptive weight generation for multi-task visual
  grounding,''
\newblock in {\em CVPR}, 2023.

\bibitem{liu2023dq}
Shilong Liu, Shijia Huang, Feng Li, Hao Zhang, Yaoyuan Liang, Hang Su, Jun Zhu,
  and Lei Zhang,
\newblock ``{DQ-DETR}: Dual query detection transformer for phrase extraction
  and grounding,''
\newblock in {\em AAAI}, 2023.

\bibitem{yu2023zero}
Seonghoon Yu, Paul~Hongsuck Seo, and Jeany Son,
\newblock ``Zero-shot referring image segmentation with global-local context
  features,''
\newblock in {\em CVPR}, 2023.

\bibitem{li2023your}
Alexander~C Li, Mihir Prabhudesai, Shivam Duggal, Ellis Brown, and Deepak
  Pathak,
\newblock ``Your diffusion model is secretly a zero-shot classifier,''
\newblock in {\em ICCV}, 2023.

\bibitem{zhao2023unleashing}
Wenliang Zhao, Yongming Rao, Zuyan Liu, Benlin Liu, Jie Zhou, and Jiwen Lu,
\newblock ``Unleashing text-to-image diffusion models for visual perception,''
\newblock in {\em ICCV}, 2023.

\bibitem{karazija2023diffusion}
Laurynas Karazija, Iro Laina, Andrea Vedaldi, and Christian Rupprecht,
\newblock ``Diffusion models for zero-shot open-vocabulary segmentation,''
\newblock {\em arXiv preprint arXiv:2306.09316}, 2023.

\bibitem{he2023discriminative}
Xuehai He, Weixi Feng, Tsu-Jui Fu, Varun Jampani, Arjun Akula, Pradyumna
  Narayana, Sugato Basu, William~Yang Wang, and Xin~Eric Wang,
\newblock ``Discriminative diffusion models as few-shot vision and language
  learners,''
\newblock {\em arXiv preprint arXiv:2305.10722}, 2023.

\bibitem{mao2016generation}
Junhua Mao, Jonathan Huang, Alexander Toshev, Oana Camburu, Alan~L Yuille, and
  Kevin Murphy,
\newblock ``Generation and comprehension of unambiguous object descriptions,''
\newblock in {\em CVPR}, 2016.

\bibitem{sohl2015deep}
Jascha Sohl-Dickstein, Eric Weiss, Niru Maheswaranathan, and Surya Ganguli,
\newblock ``Deep unsupervised learning using nonequilibrium thermodynamics,''
\newblock in {\em ICML}, 2015.

\bibitem{ho2020denoising}
Jonathan Ho, Ajay Jain, and Pieter Abbeel,
\newblock ``Denoising diffusion probabilistic models,''
\newblock in {\em NeurIPS}, 2020.

\bibitem{ronneberger2015u}
Olaf Ronneberger, Philipp Fischer, and Thomas Brox,
\newblock ``{U-Net}: Convolutional networks for biomedical image
  segmentation,''
\newblock in {\em MICCAI}, 2015.

\bibitem{yao2021cpt}
Yuan Yao, Ao~Zhang, Zhengyan Zhang, Zhiyuan Liu, Tat-Seng Chua, and Maosong
  Sun,
\newblock ``{CPT}: Colorful prompt tuning for pre-trained vision-language
  models,''
\newblock {\em arXiv preprint arXiv:2109.11797}, 2021.

\bibitem{ren2015faster}
Shaoqing Ren, Kaiming He, Ross Girshick, and Jian Sun,
\newblock ``{Faster R-CNN}: Towards real-time object detection with region
  proposal networks,''
\newblock in {\em NeurIPS}, 2015.

\bibitem{honnibal2015improved}
Matthew Honnibal and Mark Johnson,
\newblock ``An improved non-monotonic transition system for dependency
  parsing,''
\newblock in {\em EMNLP}, 2015.

\end{thebibliography}

\end{document}